\documentclass[10pt,twocolumn,letterpaper]{article}

\usepackage{cvpr}              %

\newcommand{\OURSSHORT}{LoSA}

\def \paravspace {-1\baselineskip}

\usepackage{array}
\newcolumntype{P}[1]{>{\centering\arraybackslash}p{#1}}

\usepackage[dvipsnames]{xcolor}

\newcommand{\thsup}{^{\text{th}}}

\definecolor{cvprblue}{rgb}{0.21,0.49,0.74}
\usepackage[pagebackref,breaklinks,colorlinks,citecolor=cvprblue]{hyperref}
\usepackage{multirow,multicol, makecell}
\def\paperID{2121} %
\def\confName{CVPR}
\def\confYear{2024}

\title{Time-, Memory- and Parameter-Efficient Visual Adaptation}

\author{
Otniel-Bogdan Mercea\textsuperscript{1, 2}\thanks{Work done during an internship at Google.}
    \qquad
    Alexey Gritsenko\textsuperscript{1}
    \qquad 
    Cordelia Schmid\textsuperscript{1}
    \qquad
    Anurag Arnab\textsuperscript{1}\thanks{Correspondence to aarnab@google.com.} \\
    \textsuperscript{1}Google \qquad \textsuperscript{2}University of T\"{u}bingen \\
}

\begin{document}
\maketitle
\begin{abstract}

As foundation models become more popular, there is a growing need to efficiently finetune them for downstream tasks.
Although numerous adaptation methods have been proposed, they are designed to be efficient only in terms of how many parameters are trained.
They, however, typically still require backpropagating gradients throughout the model, meaning that their training-time and -memory cost does not reduce as significantly.

We propose an adaptation method which does not backpropagate gradients through the backbone.
We achieve this by designing a lightweight network in parallel that operates on features from the frozen, pretrained backbone.
As a result, our method is efficient not only in terms of parameters, but also in training-time and memory usage.
Our approach achieves state-of-the-art accuracy-parameter trade-offs on the popular VTAB benchmark, and we further show how we outperform prior works with respect to training-time and -memory usage too.
We further demonstrate the training efficiency and scalability of our method by adapting a vision transformer backbone of 4 billion parameters for the computationally demanding task of video classification, without any intricate model parallelism.
Here, we outperform a prior adaptor-based method which could only scale to a 1 billion parameter backbone, or fully-finetuning a smaller backbone, with the same GPU and less training time.

\end{abstract}
\vspace{-0.5\baselineskip}
\section{Introduction}

\begin{figure}[t]
    \centering
    \vspace{-\baselineskip}
    \includegraphics[width=0.95\linewidth]{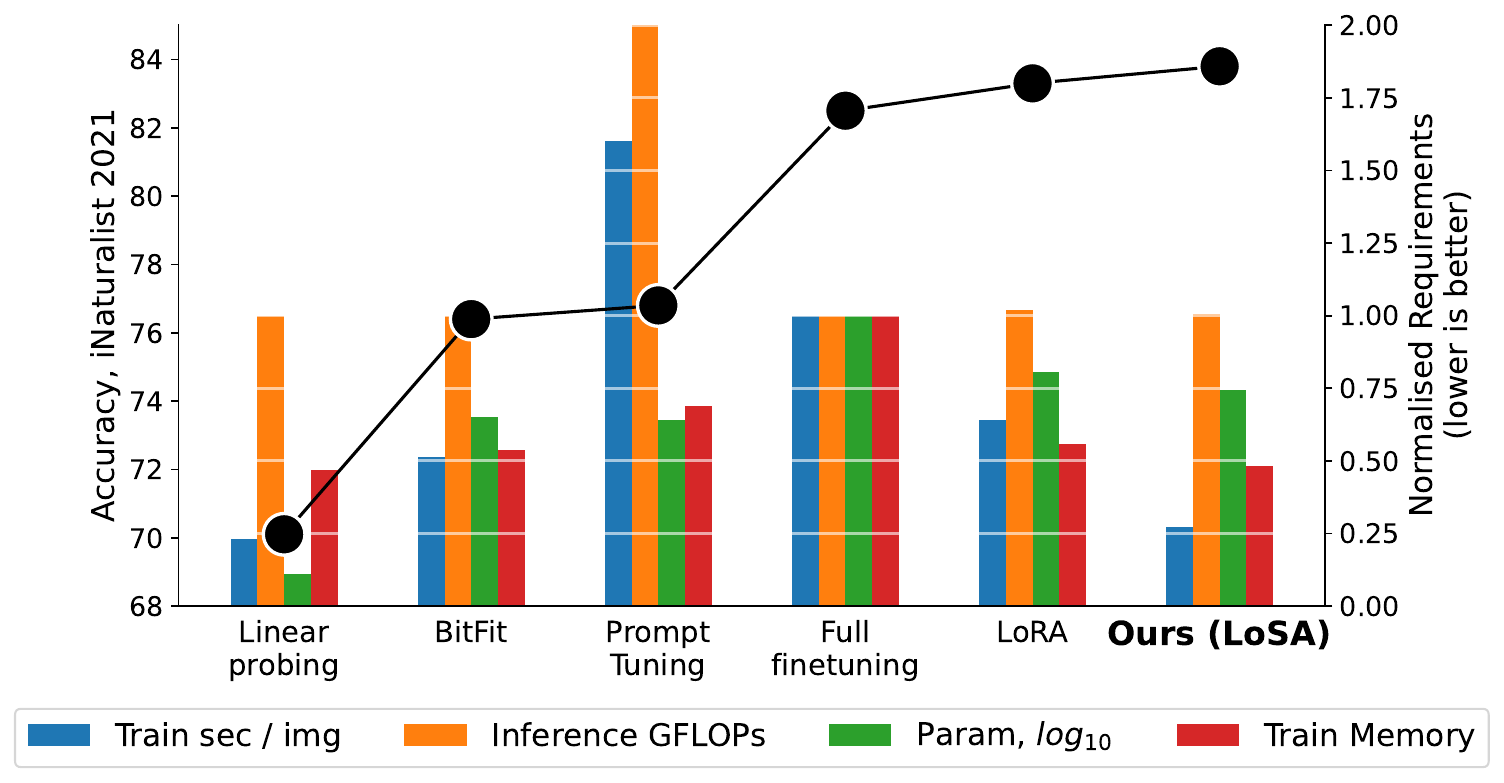}
    \vspace{-0.5\baselineskip}
    \caption{
    Parameter-efficient adaptation methods proposed in the literature are not necessarily efficient in terms of other efficiency metrics.
    Prompt-tuning~\cite{jia2022visual}, for example, learns only a few learnable prompt tokens, but is in fact slower than fully finetuning the network due to the added tokens.
    LoRA~\cite{hu2021lora} and BitFit~\cite{zaken2021bitfit}, do not substantially reduce training time as they still need to backpropagate through the whole network.
    Our method, Low-Rank Side Adaptation (LoSA), in contrast, achieves improvements across multiple efficiency metrics and tasks.
    Experiments conducted by adapting ViT-g with 1 billion parameters. %
    }
    \label{fig:teaser}
    \vspace{-1.5\baselineskip}
\end{figure}

Foundation models~\cite{brown_gpt3_neurips_2020, liu2023visual, chen2022pali, dehghani2023scaling, li2023blip} are becoming the \textit{de facto} tools of modern vision systems:
Large models, trained on massive datasets, have diverse abilities across a range of applications. %
Such foundation models are typically generalists that perform well in zero- or few-shot settings across a range of tasks~\cite{alayrac2022flamingo, brown_gpt3_neurips_2020}.
However, they typically achieve their best results on individual tasks when finetuned specifically for it, particularly when there is a large domain gap to the web-sourced pretraining data~\cite{alayrac2022flamingo, chen2023pali, li2023blip}.
As these models continue growing, it is crucial to be able to efficiently adapt, or partially finetune, large pretrained models for downstream tasks.
And particulary so for safety-critical tasks where the highest possible accuracy is critical.

Numerous efficient adaptation methods for large, pretrained models have been proposed in the literature, including LoRA~\cite{hu2021lora}, adapters~\cite{houlsby2019parameter, rebuffi2017learning} and prompt-tuning~\cite{lester2021power, li2021prefix, jia2022visual} among others.
These methods are typically designed to be parameter-efficient, or in other words, learn the smallest number of parameters to achieve high task accuracy.
However, there are numerous other relevant efficiency metrics such as training time and memory usage that are crucial, as they determine whether the adaptation method is feasible or not for larger models or low computational resources.
As shown in Fig.~\ref{fig:teaser}, although many prior methods dramatically reduce the number of learned parameters, they typically do not improve other efficiency metrics like training-time and memory as significantly.
This fact is because even though they only train a small number of parameters, they still require backpropagating gradients throughout the entire backbone of the pre-trained model.
Prompt-tuning~\cite{lester2021power, li2021prefix, jia2022visual} in particular, also adds additional tokens to the network, and therefore increases the training- and inference-time significantly as well, to the point that it may be even slower than fully finetuning the network as shown in Fig.~\ref{fig:teaser}.

More broadly, efficiency for machine learning models is a multifacted topic, as it can be measured in terms of numerous cost metrics which ultimately depend on the use case~\cite{dehghani2021efficiency, kaddour2023no}.
Nevertheless, most existing work on adapting foundation models focus only on parameter-efficiency.
In this work, we propose a method, Low-Rank Side Adaptation (LoSA), that is efficient not only in terms of the number of learned parameters, but also in the time- and accelerator-memory required during training.
This is motivated by the fact that these are key considerations for efficiently adapting large-scale models in practice, as the training costs determine whether the adaptation method is feasible in the first place.
Our work does not focus on the inference costs of such models though, which is bounded primarily by the size of the pretrained model that we are adapting.

Our method does not require backpropagating gradients through the backbone, as we keep the entire backbone frozen, and instead learn a parallel network which refines these frozen backbone features for the target task.
We find that the architecture of our parallel network is important to achieve good accuracy-efficiency trade-offs.
In particular, low-rank MLP projections which alternate between the channel, spatial and temporal (if present) dimensions achieve state-of-the-art trade-offs between accuracy, learned parameters, training-time and memory (Fig.~\ref{fig:teaser}).
And notably, unlike prior work, we thoroughly evaluate both our method and baselines on multiple efficiency indicators, not just the number of learned parameters.

Our approach achieves state-of-the-art accuracy-parameter trade-offs on the VTAB benchmark studied by most prior work in visual adaptation~\cite{jia2022visual, chen2022adaptformer, jie2022convolutional, lian2022scaling}, using the same setting of adapting ViT-Base~\cite{dosovitskiy_iclr_2021} with 86 million parameters.
However, as we believe that the goal of efficient adaptation is ultimately to adapt large-scale, pretrained foundation models, we demonstrate how our method does indeed scale to large model sizes and datasets where efficiency is paramount.
Specifically, we show how we can efficiently adapt ViT-e~\cite{chen2022pali}, a vision transformer with 4 billion parameters, on the demanding video classification task on a V100 GPU without employing any intricate model parallelism.
Moreover, we outperform prior works based on fully-finetuning smaller backbones, and a previous adaptation approach~\cite{pan2022st} for video which only scales to a 1 billion parameter backbone with the same computational budget, all whilst also training significantly faster.

In summary, we propose Low-Rank Side Adaptation (LoSA), an adaptation method which is not only parameter-efficient, but also training-time and -memory efficient. %
We can scale our method to larger backbones than previously possible (\ie ViViT-e with 4 billion parameters for video classification).
We benchmark our and prior methods on multiple efficiency indicators beyond parameters, showing that we achieve superior accuracy-efficiency trade-offs across learned parameters, training-time and -memory.
To further research in this area, we will release code for our method, and related baselines, upon acceptance.

\section{Related Work}

\begin{figure*}[t]
	\centering
        \vspace{-1.5\baselineskip}
	\includegraphics[width=0.98\linewidth, trim={0 0 90 0}, clip]{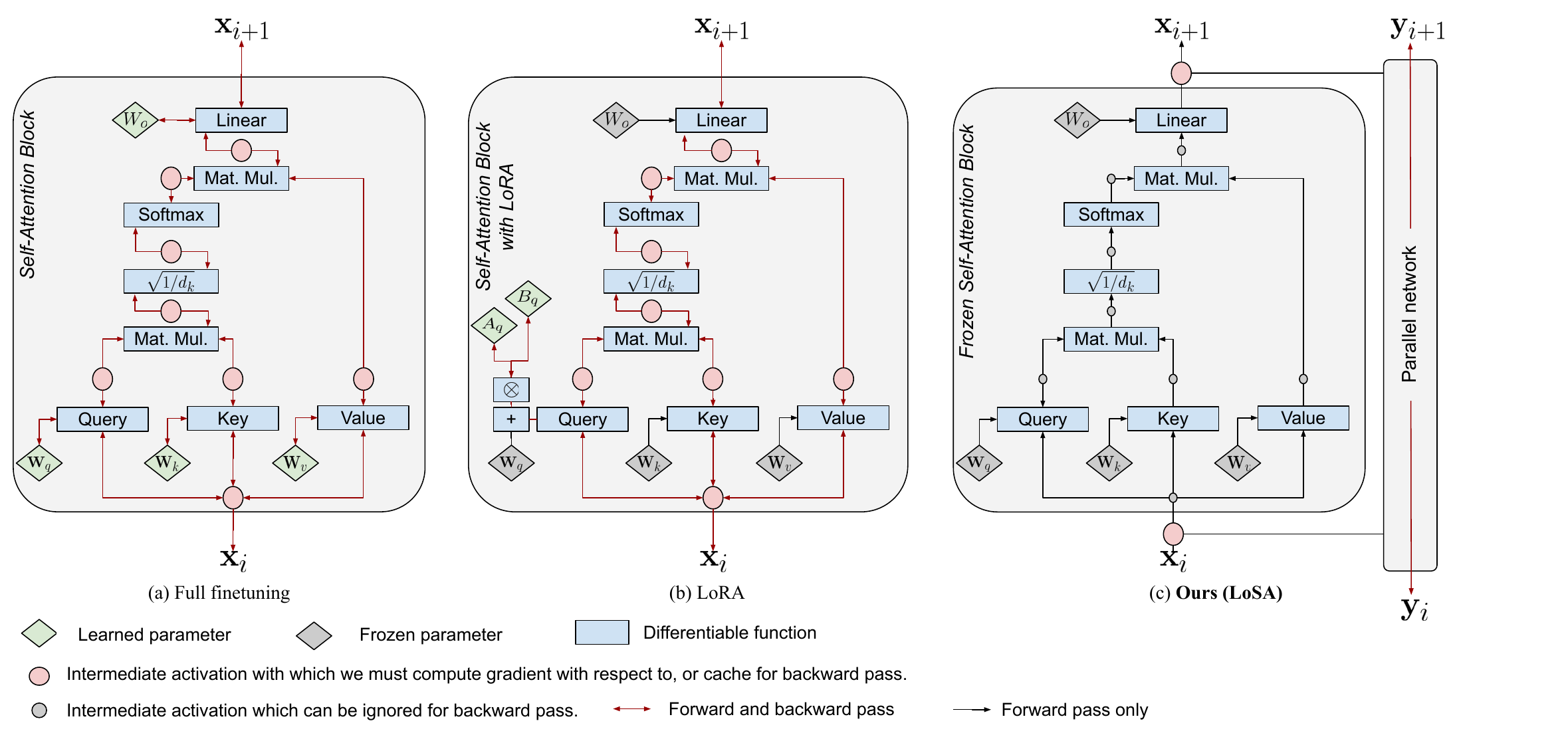} %
	\vspace{-1.25\baselineskip}
        \caption{Computational graphs of Self-Attention (SA) for the backward pass in the backbone network when performing (a) Full finetuning which requires caching or recomputing gradients with respect to large activation tensors (red ovals), which is memory- and compute-intensive (b) LoRA~\cite{hu2021lora} and (c) our proposed LoSA.
        Although LoRA (b) trains only a small number of parameters per SA block -- namely $A_q$ and $B_q$, it still requires backpropagating gradients throughout the entire backbone. Thus the computational graph is quite similar to full finetuning ($\otimes$ denotes the multiplication of two-low rank matrices).
        Our method (c), in contrast, freezes the backbone completely and does not need to backpropagate through it at all, which results in significant reductions in training time and memory.
        }

    		\label{fig:model_backprop}
        \vspace{-\baselineskip}
\end{figure*}

As large, pretrained models have become more prevalent, there has been a growing literature in efficient methods to adapt them to downstream tasks~\cite{ding2022delta, lialin2023scaling}.
The vast majority of these approaches are parameter-efficient finetuning methods which train only a small fraction of parameters in the pretrained model.
Parameter-efficient finetuning (PEFT) methods can broadly be categorised into ``additive'' methods, which add a few, new parameters to pretrained model where the original weights are frozen, and ``selective'' methods which finetune a small subset of the original network's weights~\cite{lialin2023scaling}.
Additive methods broadly consist of adapters~\cite{rebuffi2017learning, houlsby2019parameter, pfeiffer2020adapterfusion, chen2022adaptformer, karimi2021compacter, lian2022scaling, pan2022st}, which insert new learnable layers and parameters into an existing network, and prompt-tuning~\cite{jia2022visual, lester2021power, li2021prefix}, which adds learnable prompt tokens to the inputs~\cite{lester2021power} or within multiple layers~\cite{li2021prefix, jia2022visual} of a transformer network.
Selective methods on the other hand, finetune specific parameters of the original network.
For example, BitFit~\cite{zaken2021bitfit} trains only the bias terms of the network, whereas other works finetune only parameters of the attention or MLP-layers~\cite{touvron2022three, he2021towards}.
Another common theme is to reparameterise learned parameters, for example as the product of low-rank matrices~\cite{hu2021lora} or as Kronecker products~\cite{edalati2022krona} to reduce the number of learned parameters.
Numerous approaches also combine the above ideas: For example, MAM Adapters~\cite{he2021towards} combine both adapters and prompt tuning, whilst UniPelt~\cite{mao2022unipelt} makes use of low-rank weight paramterisation too.

Although the aforementioned approaches are designed with parameter-efficiency in mind, they are not necessarily computationally cheap to train.
As these methods either insert or train parameters within the original pretrained model, in order to train them, we still need to backpropagate gradients through the original network (Fig.~\ref{fig:model_backprop}).
This incurs substantial floating point operations (FLOPs) in the backward pass.
Moreover, to efficiently compute gradients in the forward pass, we typically need to cache activations during the forward pass~\cite{goodfellow2016deep, johnson:2014-thesis}.
Prompt-tuning~\cite{li2021prefix, jia2022visual} on the other hand adds even further computational overhead during both training and inference, as it inserts new tokens and thus increases the sequence length of the transformers.
Our proposed method, in contrast, keeps the entire backbone frozen and learns a parallel network that operates on activations from this backbone.
As a result, we do not need to backpropagate through the backbone, and therefore save significant computation during training.

We note that parameter-efficient methods that backpropagate gradients through the backbone, do still save memory over full-finetuning, as only a fraction of the optimiser state needs to be maintained.
For example, Adam~\cite{kingma2014adam} maintain two momentum terms per parameter, and these do not need to be kept for any of the frozen parameters in the backbone during adaptation.
As our method is still parameter-efficient, we also benefit from this property, and achieve further memory reduction during training as we need to cache fewer activations from the backbone for the backward pass.

We note that some prior works have also developed training time- and memory-efficient adaptation algorithms based on training parallel networks on top of frozen backbone features in natural language~\cite{sung2022lst, liu2022tuning, cheng2021learn}.
$\mathcal{Y}$-Tuning~\cite{liu2022tuning} and LeTS~\cite{cheng2021learn} obtained substantial time and memory reductions using this technique,
However, their methods were developed for natural language tasks, and do not apply directly to the vision problems studied in this paper, as~\cite{liu2022tuning} is based on adapting the textual label space, and~\cite{cheng2021learn} relies on bidirectional LSTMs~\cite{huang2015bidirectional} which are not typically used in vision.
LST~\cite{sung2022lst} is applicable to vision tasks. However, their results were not competitive in terms of accuracy-to-parameters trade-offs compared to existing PEFT methods. %
We carefully develop our network architecture such that we outperform prior approaches not only in terms of training-time and memory but also in terms of learned parameters.

Finally, we note that evaluating the efficiency of machine learning models is a complex subject~\cite{dehghani2021efficiency, kaddour2023no, peng2023efficiency, hooker2021hardware}.
There are numerous efficiency indicators (\eg the number of parameters, FLOPs, runtime, memory consumption), and there is often poor correlation between metrics among different techniques (for example prompt-tuning is parameter-efficient, but not efficient in terms of the other aforementioned metrics).
As selective reporting of metrics can obscure the true efficiency of an approach~\cite{dehghani2021efficiency, kaddour2023no, dahl2023benchmarking}, we strive to be as thorough as possible when measuring the efficiency of our approach and baselines, and thus go beyond reporting only parameter counts as done by prior works in this field.

\section{Proposed Approach}
\label{sec:method}

\subsection{Background}

Consider a neural network with $L$ layers, where the output of the $i\thsup$ layer, $\mathbf{x}_i$, is a function of the previous output, $\mathbf{x}_{i-1}$, and learned parameters, $\theta_i$:
\begin{equation}
	\mathbf{x}_i = f_i(\mathbf{x}_{i - 1}, \theta_i).
\end{equation}
The input to the network is $\mathbf{x}_0$.
The standard way to learn the network parameters, $\mathbf{\theta} = \{\theta_1, \ldots, \theta_L\}$ is to minimise a scalar-valued loss, $\mathcal{L}$, applied to the network's output via optimisers based on stochastic gradient descent.
Concretely, the gradient for parameter, $\theta_i$ is computed via the chain rule during backpropagation as follows:
\begin{equation}
	\frac{\partial \mathcal{L}}{\partial \theta_i} = \underbrace{\frac{\partial \mathcal{L}}{\partial \mathbf{x}_L} \frac{\partial \mathbf{x}_L}{\partial \mathbf{x}_{L-1}} \ldots \frac{\partial \mathbf{x}_{i+1}}{\partial \mathbf{x}_i}}_{\text{gradient w.r.t. previous activations}}  \frac{\partial \mathbf{x}_i}{\partial \theta_i}  
    = \frac{\partial \mathcal{L}}{\partial \mathbf{x}_i} \frac{\partial \mathbf{x}_i}{\partial \theta_i}. \label{eq:backprop}
\end{equation}

By applying the chain rule, we observe that the gradient with respect to a parameter $\theta_i$ also depends on the gradient with respect to the activations from subsequent layers (note that this analysis can easily be extended to arbitrary direct acyclic graphs~\cite{goodfellow2016deep, johnson:2014-thesis}).
In order to compute these gradients, we need to perform a ``backward pass'' which has similar computational complexity as the ``forward pass'' when computing the loss, $\mathcal{L}$, and typically requires caching the outputs of each operation, or network activations, $\mathbf{x}_i$, for efficient calculation~\cite{goodfellow2016deep, johnson:2014-thesis}.
Note that in practice, a neural network layer, such as self-attention in fact contains multiple operations within its computational graph.
For example, as shown in Fig.~\ref{fig:model_backprop}a, a self-attention layer requires computing 8 intermediate gradient terms with respect to activations, $\frac{\partial \mathcal{L}}{\partial \mathbf{x}_i}$, in order to compute gradients with respect to its query and value parameters.

\vspace{\paravspace}
\paragraph{Why parameter efficient is not always sufficient}
When training, there is substantial computational overhead in terms of the floating point operations (FLOPs) required to compute the gradient terms, $\frac{\partial \mathcal{L}}{\partial \mathbf{x}_i}$, as well as memory overhead from the activations that typically need to be cached during the forward pass %
in order to compute them. 
Parameter-efficient tuning method such as adaptors~\cite{houlsby2019parameter}, prompt-tuning~\cite{jia2022visual} and LoRA~\cite{hu2021lora} learn only a few parameters, meaning that computation can be saved by not having to compute $\frac{\partial \mathcal{L}}{\partial \theta_i}$ for the $\theta_i \in \theta$ which are frozen.
However, the majority of the computation in the backward pass is in fact from computing gradients with respect to activations, $\frac{\partial \mathcal{L}}{\partial \mathbf{x}_i}$, which must still be computed as (Fig.~\ref{fig:model_backprop}).

For example, prompt-tuning~\cite{jia2022visual}, which adds learnable prompt tokens at the start of the network, still requires computing $\frac{\partial \mathcal{L}}{\partial \mathbf{x}_i}$ with respect to each operation in the backbone, meaning that it still consumes significant amounts of FLOPs and memory during training.
A similar analysis holds for applying LoRA or other adaptors~\cite{houlsby2019parameter} at each layer of the network.
We can, however, substantially reduce the compute cost of these adaptors by only applying them to the last $k$ layers of the network, since we do not need to cache activations, or perform backpropagation, before these layers.

Based on our observation that the FLOP requirements (and therefore training time) and memory requirements of most parameter-efficient finetuning methods is still quite substantial as they must backpropagate through the model being adapted, we instead propose an alternate approach (Figs.~\ref{fig:model_backprop} and~\ref{fig:model_overview}) based on learning a \emph{parallel} network on frozen activations from the backbone model next. %

\subsection{Low-rank Side Adaptation}
\label{sec:method_losa}

\begin{figure}[t]
    \centering
    \vspace{-0.5\baselineskip}
    \includegraphics[width=0.34\textwidth,height=0.30\textheight]{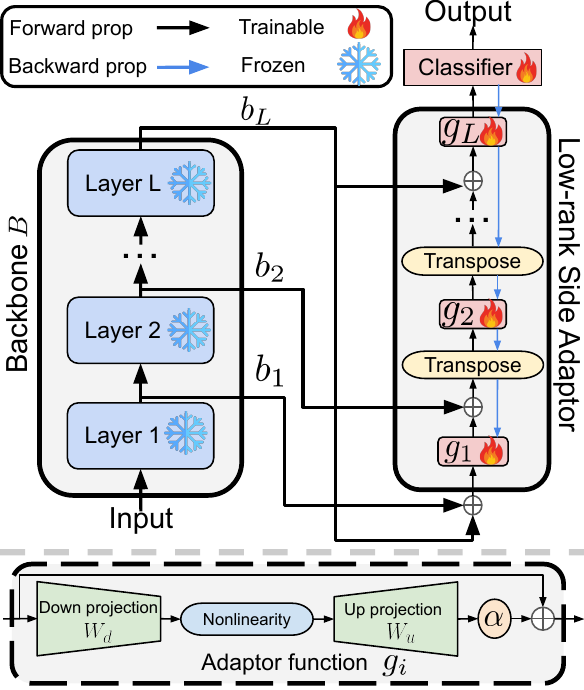}
    \caption{Overview of our approach. Top: We learn a parallel side network which iteratively refines the features obtained from a frozen backbone, $B$.
    Bottom: Our adaptation function consists of low-rank mixer modules, which allow for achieving high accuracy on a wide range of downstream tasks without sacrificing efficiency.
    }
    \label{fig:model_overview}
    \vspace{-1\baselineskip}
\end{figure}

Motivated by our observations from the previous section, we design a \emph{parallel} network that does not require backpropagating gradients through the backbone in order to train it.
Our proposed method can be thought of as a lightweight, parallel network to the main backbone, $B$ (typically a large pretrained transformer), that refines the backbone output to produce more accurate representations.

Concretely, given a neural network backbone, $B$, consisting of $L$ layers, and therefore $L$ intermediate outputs, $\mathbf{b}_1, \mathbf{b}_2, \ldots \mathbf{b}_L$,
each consisting of $n$ tokens with a hidden dimensionality of $d$, $\mathbf{b}_i \in \mathbb{R}^{n \times d}$, we learn parallel adaptor functions, $g$, which operate on these intermediate outputs to refine them.
We denote the activations of our parallel adaptor as $\mathbf{y}_i$ where $i$ denotes the layer index, and
\begin{equation}
	\mathbf{y}_i = g_i(\mathbf{b}_{i} + \mathbf{y}_{i - 1}) + \mathbf{y}_{i - 1}.
\end{equation}
As shown in Fig.~\ref{fig:model_overview}, %
we freeze the backbone, $B$, and do not backpropagate gradients to it.
Moreover, the initial input to our adaptor, $\mathbf{y}_0$ is in fact the output of the backbone, $\mathbf{b}_L$.
Together with the residual connection to the output of the previous layer, this ensures that when we train our model, our adaptor acts as an identity function with the original backbone output being retained.
The rationale is that we view our parallel adaptor, $g$, as a function that refines the features representations from the original backbone, using the original backbone features to do so. %

We design our adaptor function, $g: \mathbb{R}^d \to \mathbb{R}^d$, to be both accurate on the downstream task and efficient (in terms of parameters and floating point operations). %
To this end,  %
$g$ is a low-rank factorisation of a projection~\cite{hu2021lora}, with a down-projection, $\mathbf{W}_d: \mathbb{R}^d \to \mathbb{R}^r$, a GeLU non-linearity~\cite{hendrycks_arxiv_2016}, and then another up-projection, $\mathbf{W}_u: \mathbb{R}^r \to \mathbb{R}^d$ where $r \ll d$.
As in~\cite{hu2021lora, he2022towards, liu2022few}, we also learn a scaling term, $\alpha$, meaning that our adaptor function can be expressed as 
\begin{equation}
	g(\mathbf{x}) = \alpha \mathbf{W}_u \text{GeLU}( \mathbf{W}_d \mathbf{x} ).
	\label{eq:adaptor_fn}
\end{equation}
Note that each parallel adaptor layers contains $2rd + 1$ parameters due to the low-rank decomposition of the weight matrices~\cite{hu2021lora}.
Using additional biases in the up- and down-projections would still only add another $r + d$ parameters.

\vspace{\paravspace}
\paragraph{Modelling token interactions}
As our adaptor function, Eq.~\ref{eq:adaptor_fn}, only consists of projections and a point-wise non-linearity, it only operates on the hidden dimension of the backbone features, and does not model interactions along the spatial and temporal (in the case of videos) axes of the input data.
Therefore, we take inspiration from MLP-Mixer~\cite{tolstikhin2021mlp}, and alternate between applying our adaptor function along the channel- and token-dimensions respectively.
We denote these operations as:
\begin{equation}
	g_i(\mathbf{x}_i) = 
	\begin{cases}
		g^{\text{token}}(\mathbf{x}_i) & \text{if} \quad~i\mod2 \neq 0 \text{,}\\
		g^{\text{channel}}(\mathbf{x}_i) & \text{otherwise}.
	\end{cases}
	\label{eq:low_rank_mixer}
\end{equation}
$i$ is the backbone layer index, thus even layers correspond to ``token mixing'' and odd layers to ``channel mixing''~\cite{tolstikhin2021mlp}.

\vspace{\paravspace}
\paragraph{Extending to video} When the input is a video, we assume that the backbone features have spatiotemporal dimensions, that is $\mathbf{b}_i \in \mathbb{R}^{n_t \cdot n_s \times d}$ where $n_s$ and $n_t$ denote the spatial- and temporal-dimensions.
In this case, we factorise the token dimension further into separate spatial- and temporal-axes, that is $\mathbf{b}_i \in \mathbb{R}^{n_t \times n_s \times d}$, and apply our adaptor function separately along each dimension before summing them up.
Therefore, our adaptor function for each even layer is
\begin{equation}
	g^{\text{token}}(\mathbf{x}_i) = g^{\text{spatial}}(\mathbf{x}_i) + g^{\text{temporal}}(\mathbf{x}_i).
\end{equation}

\subsection{Discussion}

The fact that we keep the entire original backbone, $B$, frozen, and train a parallel subnetwork, means that the storage requirements of our adapted models is small, as we only need to store the parameters of our side network %
for each task.
And moreover, this makes deploying numerous backbones adapted from the same backbone, $B$, simple and efficient in terms of storage.
Other adaptor-based methods~\cite{hu2021lora, houlsby2019parameter, rebuffi2017learning} share this desirable property, although they require backpropagating through the backbone to train the model, and hence are not as efficient during training as our proposed method.
Moreover, in contrast to these prior works, our method does not require changing the internal model architecture of the backbone at all, which makes its practical implementation straightforward.

We note that some prior works have trained lightweight networks in parallel to frozen backbones for efficient adaptation in natural language processing~\cite{sung2022lst, liu2022tuning, cheng2021learn}.
Ladder Side Tuning (LST)~\cite{sung2022lst} is the most related to our approach.
However, LST was not competitive in terms of accuracy-vs-parameter trade-offs to approaches such as LoRA~\cite{hu2021lora}.
We show experimentally in Sec.~\ref{sec:exp_images} and~\ref{sec:exp_ablation} that the architectural improvements that we make over LST substantially improve the accuracy and efficiency of our approach, such that it outperforms prior works on accuracy-vs-parameter trade-offs, and substantially outperforms them in terms of training-time and -memory.
Our improvements over LST include using our low-rank mixer, instead of a regular transformer as used by LST and using the output of the backbone as the input to the parallel network, $g$.
The fact that we use our proposed low-rank mixer means that our parallel network $g$ can efficiently operate at the same hidden dimension, $d$, as the backbone, and does not require downprojecting backbone activations to the latent space of the transformer as in LST~\cite{sung2022lst}, saving compute.

Finally, we note that concurrent work, HST~\cite{lin2023hierarchical}, 
used a side network for dense prediction.
However, they use transformer layers in the side network and also employ prompt tuning in the backbone, meaning that they have to backpropagate the gradients through both the side network and the backbone.
Therefore, HST~\cite{lin2023hierarchical} does not provide benefits in terms of training-memory and speed like our method.

\section{Experimental Evaluation}

After describing our experimental setup, we present experiments on image classification in Sec.~\ref{sec:exp_images}, video classification in Sec.~\ref{sec:exp_video}, and then conclude by evaluating each of our modelling choices in Sec.~\ref{sec:exp_ablation}.

\subsection{Experimental Setup}

\vspace{-0.2\baselineskip}
\paragraph{Pretrained models}
We experimentally evaluate our method by adapting vision transformers of various backbone sizes for both image and video classification.
We note that ViT-Base~\cite{dosovitskiy_iclr_2021} (consisting of 86 million parameters) pretrained on ImageNet-21K~\cite{deng_cvpr_2009} is the most commonly used backbone across prior works~\cite{jia2022visual, chen2022adaptformer, jie2022convolutional, lian2022scaling}.
However, as a motivation of efficient adaptation is to adapt large, pretrained models, we also consider much larger vision transformers, namely ViT-H (632 million parameters), ViT-g (1 billion), ViT-G (1.8 billion) and ViT-e (4 billion)~\cite{zhai2022scaling, chen2022pali}.

\begin{table*}[t]
\vspace{-1.5\baselineskip}

\centering
\setlength{\tabcolsep}{0.3pt}
\setlength{\tabcolsep}{1pt}
\scalebox{0.82}{
\begin{tabular}{p{2.9cm} P{0.69cm}|P{0.69cm}P{0.69cm}P{0.69cm}P{0.69cm}P{0.69cm}P{0.69cm}P{0.69cm}|P{0.69cm}P{0.69cm}P{0.69cm}P{0.69cm}|P{0.69cm}P{0.69cm}P{0.69cm}P{0.69cm}P{0.69cm}P{0.69cm}P{0.69cm}P{0.69cm}|P{0.69cm}P{0.69cm}P{0.69cm}P{0.69cm}}
	\toprule[1.5pt]
	\multicolumn{2}{c|}{}&\multicolumn{7}{c|}{\textbf{Natural}}&\multicolumn{4}{c|}{\textbf{Specialized}}&\multicolumn{8}{c|}{\textbf{Structured}}&\\
	&\multicolumn{1}{c|}{\rotatebox[origin=l]{90}{Param $\downarrow$ ($10^6$)}}
	&\multicolumn{1}{c}{\rotatebox[origin=l]{90}{Cifar100}}
	&\multicolumn{1}{c}{\rotatebox[origin=l]{90}{Caltech101}}
	&\multicolumn{1}{c}{\rotatebox[origin=l]{90}{DTD}}
	&\multicolumn{1}{c}{\rotatebox[origin=l]{90}{Flower102}}
	&\multicolumn{1}{c}{\rotatebox[origin=l]{90}{Pets}}
	&\multicolumn{1}{c}{\rotatebox[origin=l]{90}{SVHN}}
	&\multicolumn{1}{c|}{\rotatebox[origin=l]{90}{Sun397}}
	&\multicolumn{1}{c}{\rotatebox[origin=l]{90}{Camelyon}}
	&\multicolumn{1}{c}{\rotatebox[origin=l]{90}{EuroSAT}}
	&\multicolumn{1}{c}{\rotatebox[origin=l]{90}{Resisc45}}
	&\multicolumn{1}{c|}{\rotatebox[origin=l]{90}{Retinopathy}}
	&\multicolumn{1}{c}{\rotatebox[origin=l]{90}{Clevr-Count}}
	&\multicolumn{1}{c}{\rotatebox[origin=l]{90}{Clevr-Dist}}
	&\multicolumn{1}{c}{\rotatebox[origin=l]{90}{DMLab}}
	&\multicolumn{1}{c}{\rotatebox[origin=l]{90}{KITTI-Dist}}
	&\multicolumn{1}{c}{\rotatebox[origin=l]{90}{dSpr-Loc}}
	&\multicolumn{1}{c}{\rotatebox[origin=l]{90}{dSpr-Ori}}
	&\multicolumn{1}{c}{\rotatebox[origin=l]{90}{sNORB-Azim}}
	&\multicolumn{1}{c|}{\rotatebox[origin=l]{90}{sNORB-Ele}}
	&\multicolumn{1}{c}{\rotatebox[origin=l]{90}{Avg Natural}}
	&\multicolumn{1}{c}{\rotatebox[origin=l]{90}{Avg Specialized}}
	&\multicolumn{1}{c}{\rotatebox[origin=l]{90}{Avg Structured}}
	&\multicolumn{1}{c}{\rotatebox[origin=l]{90}{Average}}\\
	\specialrule{0em}{1pt}{1pt}
	\hline
	\specialrule{0em}{1pt}{1pt}
	\multicolumn{22}{l}{\emph{Traditional Finetuning}}\\
	\hline
	\specialrule{0em}{1pt}{1pt}
	Full\cite{jia2022visual,jie2022convolutional} &85.8&68.9&87.7&64.3&97.2&86.9&87.4&38.8&79.7&95.7&84.2&73.9&56.3&58.6&41.7&65.5&57.5&46.7&25.7&29.1&75.9&83.4&47.6&68.9 \\
	Linear\cite{jia2022visual,jie2022convolutional}&0&64.4&85.0&63.2&97.0&86.3&36.6&51.0&78.5&87.5&68.5&74.0&34.3&30.6&33.2&55.4&12.5&20.0&9.6&19.2&69.1&77.1&26.9&57.6\\
	\hline
	\specialrule{0em}{1pt}{1pt}
	\multicolumn{22}{l}{\emph{Efficient adaptation methods}}\\
	\hline
	\specialrule{0em}{1pt}{1pt}
	BitFit\cite{zaken2021bitfit,jie2023fact}&0.10&72.8&87.0&59.2&97.5&85.3&59.9&51.4&78.7&91.6&72.9&69.8&61.5&55.6&32.4&55.9&66.6&40.0&15.7&25.1&73.3&78.3&44.1&65.2\\
	
	VPT\cite{jia2022visual,jie2022convolutional}&0.53&78.8&90.8&65.8&98.0&88.3&78.1&49.6&81.8&96.1&83.4&68.4&68.5&60.0&46.5&72.8&73.6&47.9&32.9&37.8&78.5&82.4&55.0&72.0 \\
    RS-Bypass.~\cite{jiang2023restuning}&0.42&64.5&88.8&73.2&99.4&90.6&63.5&\underline{57.2}&85.5&95.2&82.4&75.2&70.4&61.0&40.2&66.8&79.2&52.6&26.0&\underline{49.3}&76.7&84.6&55.7&72.3 \\

    Adapter \cite{houlsby2019parameter, jie2022convolutional}&0.16&69.2&90.1&68.0&98.8&89.9&82.8&54.3&84.0&94.9&81.9&75.5&80.9&65.3&48.6&78.3&74.8&48.5&29.9&41.6&79.0&84.1&58.5&73.9 \\
	LoRA\cite{hu2021lora,jie2022convolutional}&0.29&67.1&91.4&69.4&98.8&90.4&85.3&54.0&84.9&95.3&84.4&73.6&\underline{82.9}&\bf69.2&49.8&78.5&75.7&47.1&31.0&44.0&79.5&84.6&59.8&74.5
	\\
	AdaptFormer\cite{chen2022adaptformer,jie2022convolutional}&0.16&70.8&91.2&70.5&99.1&90.9&86.6&54.8&83.0&95.8&84.4&76.3&81.9&64.3&49.3&80.3&76.3&45.7&31.7&41.1&80.6&84.9&58.8&74.7 \\
    NOAH \cite{zhang2022neural, jie2022convolutional}&0.36&69.6&92.7&70.2&99.1&90.4&86.1&53.7&84.4&95.4&83.9&75.8&82.8&\underline{68.9}&49.9&81.7&81.8&48.3&32.8&44.2&80.3&84.9&61.3&75.5\\
    FacT-TK \cite{jie2023fact}& \underline{0.06}&70.6&90.6&70.8&99.1&90.7&88.6&54.1&84.8&96.2&84.5&75.7&82.6&68.2&49.8&80.7&80.8&47.4&33.2&43.0&80.6&85.3&60.7&75.6\\
	
    RS~\cite{jiang2023restuning}&0.55&75.2&92.7&71.9&99.3&\bf91.9&86.7&\bf{58.5}&86.7&95.6&85.0&74.6&80.2&63.6&50.6&80.2&85.4&55.7&31.9&42.0&82.3&85.5&61.2&76.3 \\	
    Convpass \cite{jie2022convolutional}&0.33&72.3&91.2&72.2&99.2&90.9&\bf91.3&54.9&84.2&96.1&85.3&75.6&82.3&67.9&\underline{51.3}&80.0&85.9&53.1&36.4&44.4&81.7&85.3&62.7&76.6\\
    HST~\cite{lin2023hierarchical}&0.78&76.7&\bf94.1&74.8&\underline{99.6}&\underline{91.1}&\underline{91.2}&52.3&\bf{87.1}&96.3&\bf88.6&\underline{76.5}&\bf85.4&63.7&\bf{52.9}&81.7&87.2&56.8&35.8&\bf52.1&\bf82.8&\bf87.1&64.5&78.1 \\
	\hline
	
	\OURSSHORT, $r = 16$&0.19&\underline{82.5}&92.8&76.1&\bf99.7&90.5&82.0&55.8&86.6&\bf97.1&87.0&\bf76.7&81.5&62.3&48.6&82.1&\bf94.2&\bf61.7&\underline{47.9}&45.6&\bf82.8&86.9&\bf65.5&\bf78.4\\ %
	\OURSSHORT, $r = 8$&0.10&82.2&92.7&\bf76.7&\bf99.7&90.7&81.0&55.4&\underline{86.9}&\bf97.1&\underline{87.4}&\underline{76.5}&79.9&61.8&48.6&\underline{82.4}&\underline{92.3}&\underline{61.1}&\bf48.7&47.3&\underline{82.6}&\underline{87.0}&\underline{65.3}&\underline{78.3}\\ %
	
	\OURSSHORT, $r = 4$&\bf{0.05}&\bf82.7&\underline{93.0}&\underline{76.2}&\bf99.7&89.8&80.0&56.1&86.3&\underline{96.7}&86.7&76.3&78.8&61.4&48.0&\bf82.6&91.7&58.4&46.9&47.6&82.5&86.5&64.4&77.8\\ %

	\bottomrule[1.5pt]

\end{tabular}
}
\vspace{-0.75\baselineskip}
\caption{
Comparison to state-of-the-art parameter-efficient finetuning methods on VTAB-1K~\cite{zhai2019large}.
Following standard practice, the final ``Average'' is the average of three preceding groupwise averages.
Parameters denotes the number of learnable parameters excluding the final classification layer, as the number of parameters in this final layer depend on the number of classes, which varies between 2 and 397.
Each variant of our model, which we obtain by varying the rank $r$ of our Low-rank Mixer Block, achieves better accuracy-parameter trade-offs than previous approaches, when using the same ViT-B backbone.
\textbf{Best results} are bolded, and \underline{second-best} underlined.
}
\label{tab:vtab}
\vspace{-1.5\baselineskip}
\end{table*}

\vspace{\paravspace}
\paragraph{Efficiency metrics}
In addition to accuracy on the final task, as our efficiency metrics, we report the training time (measured in the number of images per second per core on a V100 GPU), training memory consumption in Gigabytes (GB), number of FLOPs during inference, and number of trainable parameters.
We find that the inference time is largely determined by the size of the pretrained backbone, and not the adaptation method (except for prompt-tuning where it increases substantially).
Therefore, we only include it in the supplement for completeness.
When reporting the number of learnable parameters, we do not report the parameters in the final classification layer for clarity. The reason being that for datasets with a large number of classes, \eg iNaturalist2021 has 10 000 classes, the number of learned parameters is dominated by the final layer, and not the architecture of the various adaptation methods. 

\vspace{\paravspace}
\paragraph{Implementation details}
We include exhaustive details of our training hyperparameters in the supplement.
% We will also release code for our method, and the various baselines that we have implemented upon acceptance.

\subsection{Experiments on Images}
\label{sec:exp_images}

The most common setting for studying parameter-efficient finetuning is on the VTAB~\cite{zhai2019large} benchmark, by adapting ViT-B pretrained on ImageNet-21K.
We first compare our method to prior works in this scenario. %
However, as a key motivation for efficient adaptation is to use large-scale models, we perform experiments on more challenging, large-scale classification datasets including iNaturalist2018~\cite{inaturalist18} and 2021~\cite{inaturalist21} and Places365~\cite{zhou2017places} 
with ViT-g.
Note that we do not consider ImageNet, as we found performance to be saturated and not representative for comparing different methods. We nevertheless include it in the supplement. %

\vspace{\paravspace}
\paragraph{VTAB benchmark}
\label{sec:exp_vtab}

VTAB~\cite{zhai2019large} is the most common benchmark in the vision community for evaluating parameter-efficient finetuning methods.
It consists of 19 datasets spanning a range of image domains.
The benchmark is designed to evaluate transfer-learning performance, and each dataset contains exactly 1000 training examples.

Prior works have evaluated on this benchmark, using a ViT-Base backbone and benchmarking efficiency through the number of trained parameters.
We compare to these works, in the same setting in Tab.~\ref{tab:vtab}.
We observe that we achieve state-of-the-art trade-offs between the accuracy and the number of trainable parameters.
We can control the number of parameters in our model, by varying $r$, the size of our hidden layer within our side network (Sec.~\ref{sec:method_losa}).
With $r = 16$, we obtain a parameter-efficient model which outperforms prior works on this benchmark using the same ViT-Base model pretrained on ImageNet-21K.
In particular, we outperform concurrent work, HST~\cite{lin2023hierarchical}, whilst using 4 times fewer parameters.
And when we decrease $r$ to $r = 4$ and $r = 8$, we continue to achieve superior parameter-accuracy trade-offs, with $r = 4$ having the lowest number of parameters and still achieving the second-highest accuracy.

Table~\ref{tab:vtab} therefore shows that our proposed LoSA method outperforms the state-of-the-art with respect to accuracy-parameter trade-offs.
Next, we analyse our method on additional efficiency indicators at a larger scale. %

\vspace{\paravspace}
\paragraph{Large-scale image classification}
\label{sec:exp_large_scale_image}

\begin{figure*}[t]
    \centering
    \vspace{-0.5\baselineskip}
    \includegraphics[width=0.99\textwidth]{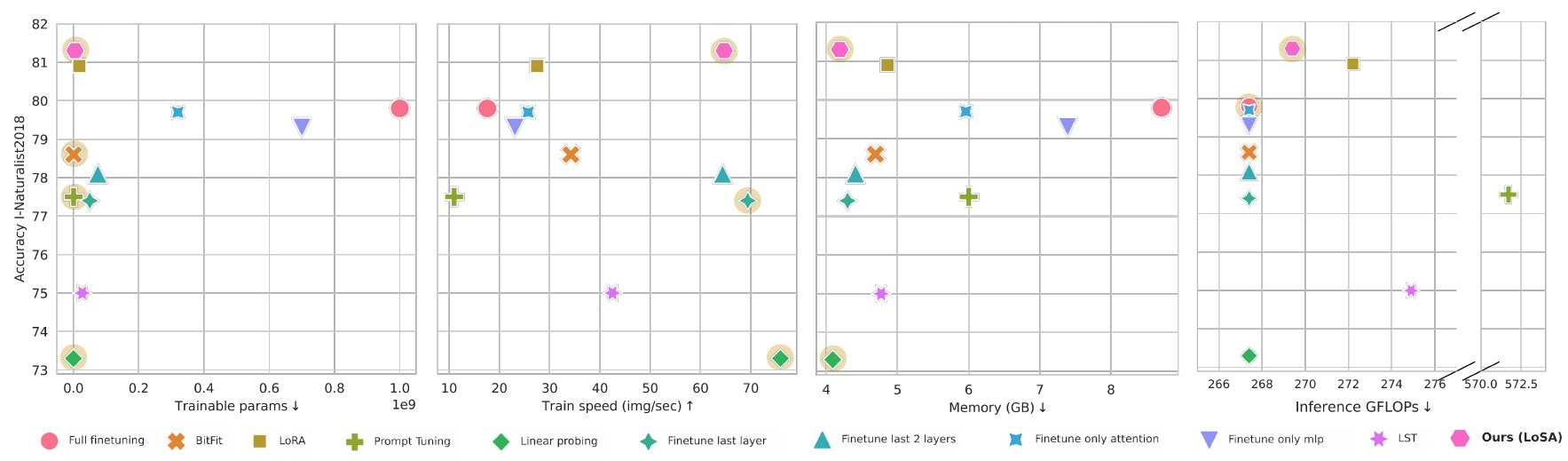}
    \vspace{-0.75\baselineskip}
    \caption{Comparison of trade-offs of accuracy with respect to learned parameters, training memory, inference GFLOPs and training speed. 
    	Our approach, LoSA, is consistently on the Pareto frontier (denoted by shaded yellow circles), as there is no method that is both more accurate and more efficient than it, across multiple efficiency metrics.
    	Results are on the iNaturalist 2018 dataset, using a ViT-g backbone with 1 billion frozen parameters.
	}
    \label{fig:results_inaturalist}
    \vspace{-\baselineskip}
\end{figure*}

We consider more challenging, larger-scale image classification datasets in the form of iNaturalist 2018~\cite{inaturalist18} (438K training examples, 8142 classes), iNaturalist 2021~\cite{inaturalist21} (2.7M training examples, 10000 classes) and Places365~\cite{zhou2017places} (1.8M training examples, 365 classes).
The iNaturalist datasets also have a long-tail, increasing their difficulty.

Figure~\ref{fig:results_inaturalist} evaluates our method and baselines across a wide range of efficiency indicators.
We compare to LoRA~\cite{hu2021lora}, BitFit~\cite{zaken2021bitfit}, prompt tuning~\cite{jia2022visual} and LST~\cite{sung2022lst} as representative efficient adaptation algorithms.
We also compare to full-finetuning, linear probing, finetuning the last one or two layers, and also finetuning only the attention or MLP layers~\cite{touvron2022three}.
Extensive implementation details of these baselines are included in the supplement.
We visualise the Pareto frontier across each of these methods with yellow, shaded circles, to show the relative trade-offs in each efficiency metric.

Consistent with our experiments on VTAB (Tab.~\ref{tab:vtab}), we show superior trade-offs in terms of accuracy and parameters.
We also outperform full-finetuning, and this suggests that large backbones are overparameterised, and can overfit when fully-finetuned.
Therefore, efficiently adapting a subset of parameters may lead to higher performance.
Numerous works across domains~\cite{pan2022st,chen2022adaptformer,sung2022lst, lin2022frozen} have also observed that efficient finetuning methods can outperform full-finetuning in certain tasks.

Figure~\ref{fig:results_inaturalist} also shows that our method uses less training memory than other baselines except linear probing.
As described in Sec.~\ref{sec:method}, these improvements stem from the fact that we do not need to backpropagate gradients through the backbone.
However, we also improve upon the accuracy of linear probing by 8 points.
We note similar observations when comparing the training speed to overall accuracy for the same reason that we do not need to backpropagate through the backbone.
Notably, we are 2.35 times faster than LoRA whilst achieving 0.4 points higher accuracy.

Overall, our method is Pareto-optimal for learned parmeters, training speed and training memory consumption, emphasising the advantages of our approach.
The inference GFLOPs is largely unchanged among the different approaches, as they are all dominated by the cost of the ViT-g backbone which is constant across them.
Notably, prompt tuning uses substantially more GFLOPs and training time as in order to achieve good accuracy, we needed to use 16 prompts in the first 24 layers of the network.

We show full tables with the complete data for Fig.~\ref{fig:results_inaturalist} in the supplement, as well as accuracies on iNaturalist 2021 (Fig.~\ref{fig:teaser}) and Places365 where we observe consistent trends.

\subsection{Experiments on Video}
\label{sec:exp_video}

\begin{table}[t]
    %\vspace{-1\baselineskip}
    \centering
    \resizebox{\linewidth}{!}{
    \begin{tabular}{lccccc}
    \toprule
    \multirowcell{2}[0pt][l]{Method} & \multirowcell{2}{Backbone} & \multirowcell{2}{Params $\downarrow$ \\ ($10^6$)} &   \multirowcell{2}{Memory $\downarrow$\\ (GB)} & \multirowcell{2} {Train $\uparrow$\\ (clip/second)}  & \multirowcell{2}{Accuracy $\uparrow$}\\  
    & &  & & &\\               
    \midrule
    Full finetuning~\cite{arnab2021vivit} & ViViT-H & 635 & 14.34  &0.61 & 84.9 \\ %
    & ViViT-g & 1039 & --  & -- & OOM  \\
    & ViViT-G & 1879 &  --  & -- & OOM \\
    & ViViT-e & 3879 &  --  & -- & OOM \\
    \midrule
    ST Adapter~\cite{pan2022st} & ViViT-H &127 & 9.17 & 0.66& 80.9\\ %
    & ViViT-g & 174 &  11.47  & 0.50 &83.9 \\ %
    & ViViT-G & 247 &  --  & -- & OOM\\  %
    & ViViT-e & 310&  --  & -- & OOM \\ %

    \midrule
    Linear Probing & ViViT-H & 0 & \textbf{2.88} & \textbf{2.47} & 75.0\\ %
    & ViViT-g & 0 &  3.62 & 1.94 & 80.7\\ %
    & ViViT-G & 0 & 5.24  & 1.37 & 81.1\\  %
    & ViViT-e & 0 & 9.05 & 0.92  & 83.1  \\ %
    \midrule
     Ours, LoSA & ViViT-H &3.61  &  3.41 & 2.33 & 80.0 \\  %
     & ViViT-g & 4.78 & 4.92  & 1.73 & 84.0  \\ %
     & ViViT-G & 6.53 & 7.72  & 1.22 & 84.4  \\ %
     & ViViT-e & 8.03 & 12.54  & 0.83 & \textbf{86.2} \\ %

    \bottomrule
    \end{tabular}
    }
    \vspace{-0.75\baselineskip}
    \caption{Video classification results on Kinetics 400, using Nvidia V100 GPUs with 16 GB of RAM.
    Our proposed method can scale to a frozen ViViT-e backbone, as it does not backpropagate gradients through the backbone.
    And thanks to the larger backbone size, can outperform fully-finetuning a smaller ViViT-H model.
    \label{tab:video_classification}
    }
    \vspace{-1.5\baselineskip}
\end{table}

We further demonstrate the scalabilty of our method by performing video classification experiments on Kinetics 400~\cite{kay_arxiv_2017}.
We use an unfactorised ViViT model~\cite{arnab2021vivit}, which allows us to use the same pretrained image checkpoints from our previous experiments.

Video classification requires processing substantially more tokens: each model in Tab.~\ref{tab:video_classification} is trained with 32 frames, a frame resolution of 224, and a tubelet size of $14 \times 14 \times 2$,  
which corresponds to 4096 tokens.
As shown in Tab.~\ref{tab:video_classification} it is only possible to fully finetune a ViViT-H model with 636 million parameters.
We also compare to ST-Adapter~\cite{pan2022st}, an existing work for adapting video models, based on the authors' public code %
as detailed in the suppement, 
and observe that it does not scale beyond ViViT-g with 1 billion parameters. %
However, our approach, as it does not backpropagate gradients through the backbone unlike the previous two approaches, can scale up to ViViT-e with 4 billion frozen parameters, using Nvidia V100 GPUs with 16GB of RAM.
Moreover, we achieve higher accuracy than either ST-Adapter or full finetuning (86.2 vs 84.9) whilst also training faster and using less memory.
We note, however, that although our model is faster to train than these two baselines, our final ViViT-e model is still slower at inference time due to its larger backbone.

Linear probing is the only baseline that allows us to scale to the ViViT-e backbone.
However, its performance is lower as well, and consistent with our previous experiments on images. %
Note that whilst it is possible in theory to use model parallelism to split a large model across multiple accelerators, we have not implemented this for any approach in Tab.~\ref{tab:video_classification} as it is not common practice. %

\subsection{Ablation Study}
\label{sec:exp_ablation}

Finally, we analyse the effect of our various modelling choices.
Unless otherwise stated, we perform experiments on iNaturalist 2018~\cite{inaturalist18} using a ViT-g backbone.

\vspace{\paravspace}
\paragraph{Adaptor function $g$}

\begin{table}[t]
    \centering
    \vspace{-0.5\baselineskip}
    \resizebox{\linewidth}{!}{
    \begin{tabular}{lcccccc}
    \toprule
    \multirowcell{2}[0pt][l]{Adaptor function, $g$} &  \multirowcell{2}{Learned $\downarrow$ \\ Params (M) } & \multirowcell{2}{Train Memory $\downarrow$ \\ (GB)  } & \multirowcell{2}{Train $\uparrow$ \\ (img/sec) }  & \multirowcell{2}{GFLOPs $\downarrow$} & \multirowcell{2}{Acc $\uparrow$}\\  
    & & & & & \\               
    \midrule
    Transformer & 27 & 4.77  &42.5& 274.9 & 75.0 \\ %
    Low-rank MLP & 7.7 & 4.21 & \bf65.7 & \bf269.4& 80.7 \\ %
    Low-rank Mixer~(\OURSSHORT) & \bf4.8&  \bf4.19& 64.8 & \bf269.4& \bf81.3\\ %
    \bottomrule
    \end{tabular}
    }
    \vspace{-0.75\baselineskip}
    \caption{Ablation of our adaptor function, $g$. LST~\cite{sung2022lst} uses a regular transformer which we find to perform the worst, whilst our Low-rank Mixer achieves the best accuracy-efficiency trade-off. 
    }
    \label{tab:ladder_function_ablation}
\end{table}

Table~\ref{tab:ladder_function_ablation} considers different alternatives for our adaptor function, $g$.
Our first baseline is to use a regular transformer, with hidden dimension $r = 48 $, as also done in LST~\cite{sung2022lst}.
However, as shown in Tab.~\ref{tab:ladder_function_ablation}, this performs the worst, both in terms of accuracy and efficiency.
Although we could use smaller hidden dimensions to improve efficiency, we assume fewer parameters would not increase accuracy, leaving a clear gap to a Low-rank MLP (Eq.~\ref{eq:adaptor_fn}).
Alternating between ``channel mixing'' and ``token mixing''~\cite{tolstikhin2021mlp} with our proposed Low-rank Mixer (Eq.~\ref{eq:low_rank_mixer}) achieves the best accuracy. 
Moreover, this also improves efficiency metrics, because for images, the number of tokens (256) is less than the hidden size ($d = 1408$) for our backbone.
Therefore, every second mixer block has fewer parameters and GFLOPs compared to the Low-rank MLP.

\vspace{\paravspace}
\paragraph{Backbone activation}

\begin{table}[t]
    \centering
    \vspace{-0.6\baselineskip}
    \resizebox{\linewidth}{!}{
    \begin{tabular}{lcccccc}
    \toprule
    \multirowcell{2}[0pt][l]{Backbone activation} &  \multirowcell{2}{Learned $\downarrow$ \\ Params (M)} & \multirowcell{2}{Train Memory$\downarrow$ \\ (GB) } & \multirowcell{2}{Train $\uparrow$ \\ (img/sec) } & \multirowcell{2}{GFLOPs $\downarrow$} & \multirowcell{2}{Acc $\uparrow$}\\  
    & &  && \\               
    \midrule
    
    MLP & 4.8 & 4.19& 64.8 & 269.4 & 77.7\\ %
    Multihead attention & 4.8 & 4.19& 64.8 & 269.4  & 79.1\\ %
    MLP and Multihead attention & 9.1 & 4.29 &51.8& 271.2  & 81.0 \\ %
    Encoder block (\OURSSHORT) & 4.8& 4.19 &  64.8&  269.4 & \bf81.3\\ %
    \bottomrule
    \end{tabular}
    }
    \vspace{-0.75\baselineskip}
    \caption{The effect of different activations extracted from the backbone. The output of the transformer encoder block performs the best as it is the sum of the MLP- and self-attention-block outputs.}
    \label{tab:ablation_feature_extraction}
    \vspace{-1\baselineskip}
\end{table}

Table~\ref{tab:ablation_feature_extraction} analyses the various activations that we can extract from our backbone to pass to the adaptor function, $g$.
We find that the final output of the transformer encoder block, performs better than the alternatives: output of the MLP, output of the Multihead Self-Attention (MHSA) block, or in fact, the combination of both.
When we use outputs of both MHSA- and MLP-layers, our side network effectively has double the depth as we apply $g$ to each output in sequence.
Using the final encoder block likely performs the best, as via the use of residual connections, it is the sum of both MHSA- and MLP-blocks~\cite{dosovitskiy_iclr_2021}, and thus contains information from both.

\vspace{\paravspace}
\paragraph{Number of layers in side network}

\begin{table}[t]
    \centering
    \vspace{-0.5\baselineskip}
    \resizebox{\linewidth}{!}{
    \begin{tabular}{lcccccc}
    \toprule
    \multirowcell{2}[0pt][l]{Layers in side network} &  \multirowcell{2}{Learned $\downarrow$\\ Params (M)} & \multirowcell{2}{Train Memory $\downarrow$\\ (GB)  } & \multirowcell{2}{Train $\uparrow$\\ (img/sec) } & \multirowcell{2}{GFLOPs $\downarrow$} & \multirowcell{2}{Acc $\uparrow$}\\  
    & & &  &  & \\               
    \midrule
    
    Linear Probing (0) & 0 &  4.09 &  75.7 & 267.4  &73.3  \\ %
    \midrule
    1 &0.184 & 4.09 & 75.2 & 267.5 & 77.8 \\  %
    2 &0.217 & 4.10 &74.8   &267.5& 78.0 \\ %
    4 & 0.431&4.10 & 74.2  &267.6& 79.3 \\ %
    8 & 0.861  &  4.11    & 73.2      &   267.8 & 80.0 \\ %
    40 (All) &4.298 &4.19 & 64.8 &269.3& 80.6 \\ %
    
    \bottomrule
    \end{tabular}
    }
    \vspace{-0.75\baselineskip}
    \caption{Effect of the number of layers in our side network. As we use a ViT-g backbone, there are 40 layers in the backbone, and we uniformly sample backbone activations to pass to our side network. No layers corresponds to linear probing, which performs substantially worse than using just a single side adaptor layer.}
    \label{tab:ablation_layers_no_maphead}
    \vspace{-0.5\baselineskip}
\end{table}

Another method of controlling the accuracy-efficiency trade-off of our proposed method is to vary the number of layers in our side network.
Instead of operating on activations from each layer of the backbone, we select $k$ evenly-spaced layers instead in Tab.~\ref{tab:ablation_layers_no_maphead}.
Using no layers from our side network results in linear probing, and we observe that adding just a single side adaptor layer to the end of the backbone improves accuracy significantly by 4.5 points. This suggests that our side network is able to learn complementary information to the backbone.
Adding further layers gives small and consistent improvements.
In our previous experiments, we used all layers from the backbone to obtain higher accuracy.

\vspace{\paravspace}
\paragraph{Side network inputs}

\begin{table}[t]
    \vspace{-0\baselineskip}
    \centering
    \resizebox{\linewidth}{!}{
    \begin{tabular}{lcccccc}
    \toprule
    \multirowcell{2}[0pt][l]{Side network input, $\mathbf{y}_0$} &  \multirowcell{2}{Learned $\downarrow$ \\ Param (M)} & \multirowcell{2}{Train Memory $\downarrow$ \\ (GB) } & \multirowcell{2}{Train $\uparrow$ \\ (img/sec)}& \multirowcell{2}{GFLOPs $\downarrow$} & \multirowcell{2}{Acc $\uparrow$}\\  
    & & &  & & \\               
    \midrule
    Backbone input & 4.8 & 4.06  & 64.8 &  269.4  & 60.9\\     %
    Backbone output (\OURSSHORT) & 4.8 & 4.06 & 64.8 & 269.4 & \textbf{61.3} \\ %
    \bottomrule
    \end{tabular}
    }
    \vspace{-0.75\baselineskip}
    \caption{Ablation of inputs to our side network on Places365. Using the backbone outputs provides a small improvement compared to using the backbone inputs, as done in LST~\cite{sung2022lst}.}
    \label{tab:ablation_input_features}
    \vspace{-1.25\baselineskip}
\end{table}

Finally, Tab.~\ref{tab:ablation_input_features} considers the effect of using either the output of the backbone, $\mathbf{b}_L$, as the input to our side network, $\mathbf{y}_0$, or rather using the input to the backbone itself (\ie tokenised image patches, $\mathbf{b}_0$). %
The interpretation of $\mathbf{b}_L$ is that the side network refines the outputs of the backbone, given further intermediate activations from the backbone.
The interpretation of $\mathbf{b}_0$ is that the side network iteratively refines backbone features of the $i^{\text{th}}$ layer given activations from layers before it, and was also used in LST~\cite{sung2022lst}.
We find that the former approach improves results slightly, and thus used it for all experiments.

\section{Conclusion and Future Work}

We have introduced Low-rank Side Adaptation (LoSA), an adaptation method that is efficient not only in terms of parameters, but also training-time and memory, since we do not have to backpropagate gradients through the backbone.
We achieve state-of-the-art results on VTAB, and show that we can scale our method, unlike previous works, to ViT-e (4 billion parameters) without model parallelism.
Future work is to extend our method to more complex vision tasks.

{
    \small
    \bibliographystyle{ieeenat_fullname}
    \bibliography{bibliography}
}

\clearpage
\appendix

\section*{Appendix}
In this appendix, we provide further experiments on large-scale image classification (Sec.~\ref{sec:Large_scale}), futher details and ablation studies of our baselines (Sec.~\ref{sec:Implementation_details}), detail our hyperparameters (Sec.~\ref{sec:Hyperparam}), and finally, include a more extensive comparison to prior work on the VTAB benchmark~\cite{zhai2019large} (Sec.~\ref{sec:VTAB}) which we could not do in the main paper due to space constraints.

\section{Further experiments on large scale image-classification} \label{sec:Large_scale}

\begin{figure*}[]
    \centering
    \vspace{-\baselineskip}
    \includegraphics[width=0.99\textwidth]{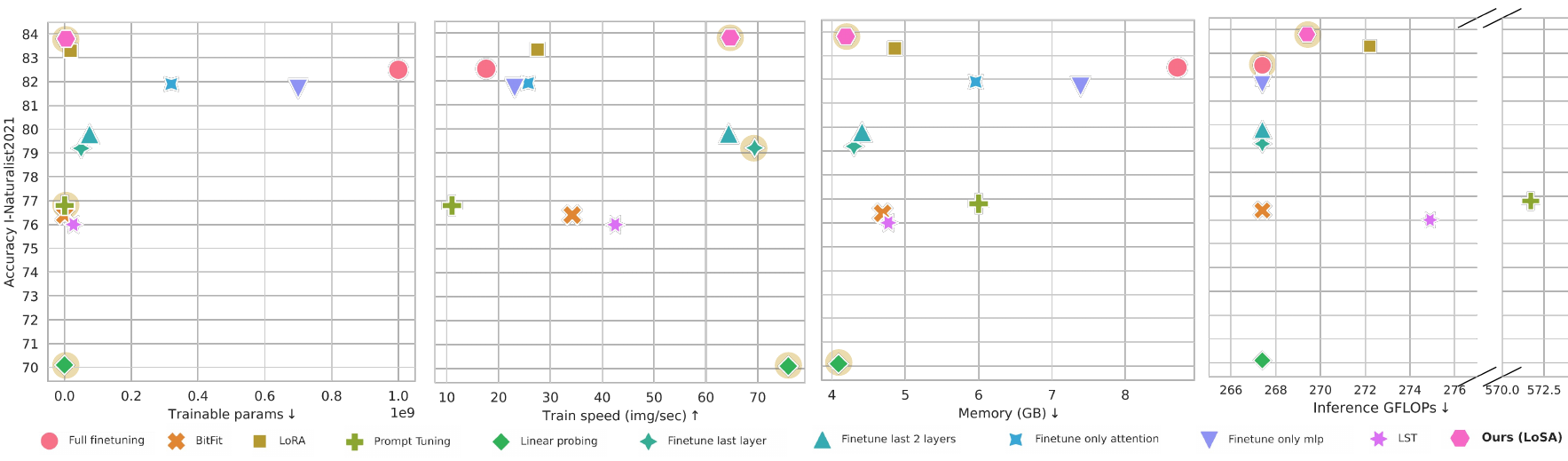}
    \vspace{-0.75\baselineskip}
    \caption{Comparison of trade-offs of accuracy with respect to learned parameters, training memory, inference GFLOPs and training speed. 
    	Our approach, LoSA, is consistently on the Pareto frontier (denoted by shaded yellow circles), as there is no method that is both more accurate and more efficient than it, across multiple efficiency metrics.
    	Results are on the iNaturalist2021 dataset, using a ViT-g backbone with 1 billion frozen parameters.
	}
    \label{fig:results_inaturalist2021}
\end{figure*}

\begin{figure*}[t]
    \centering
    \vspace{-\baselineskip}
    \includegraphics[width=0.99\textwidth]{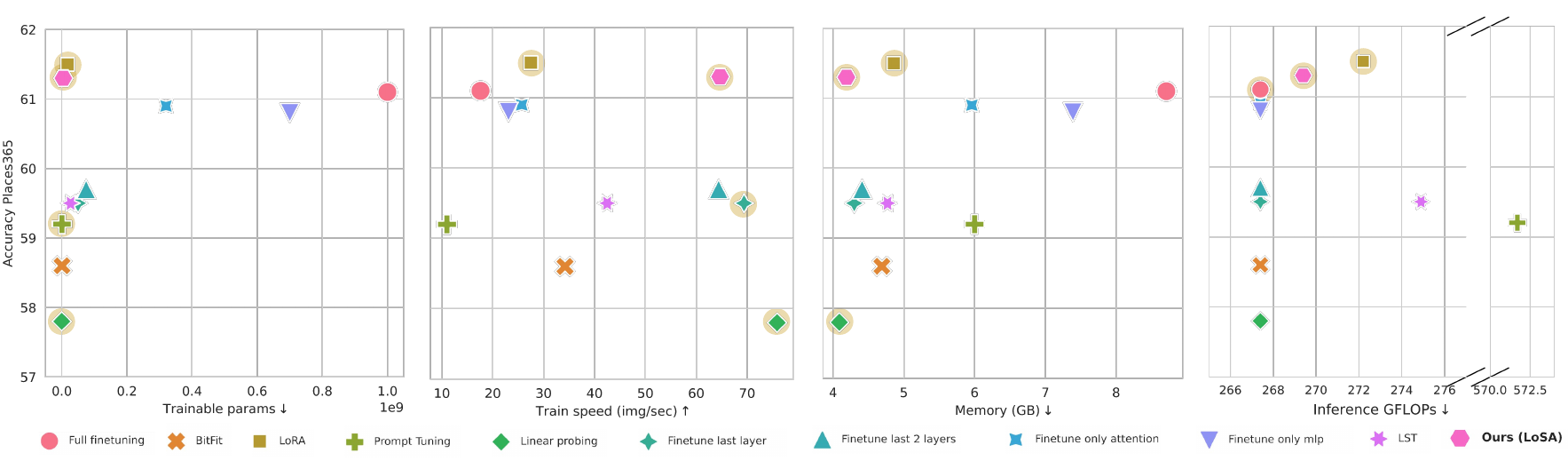}
    \vspace{-0.75\baselineskip}
    \caption{Comparison of trade-offs of accuracy with respect to learned parameters, training memory, inference GFLOPs and training speed. 
    	Our approach, LoSA, is consistently on the Pareto frontier (denoted by shaded yellow circles), as there is no method that is both more accurate and more efficient than it, across multiple efficiency metrics.
    	Results are on the Places365 dataset, using a ViT-g backbone with 1 billion frozen parameters.
	}
    \label{fig:results_places365}
    \vspace{-\baselineskip}
\end{figure*}

\begin{table*}[t]
    \centering
    \resizebox{\linewidth}{!}{
    \begin{tabular}{lccccccccccc}
     \toprule
     \multirowcell{2}{Method} &  \multirowcell{2}{Update $\downarrow$\\ Param (M)} & \multicolumn{2}{c}{Memory Usage (GB)} & \multicolumn{2}{c} {Speed img/sec} & \multirowcell{2}{GFlops $\downarrow$} & \multirowcell{2}{Acc $\uparrow$\\ Places365}& \multirowcell{2}{Acc $\uparrow$ \\ iNaturalist2018}& \multirowcell{2}{Acc $\uparrow$ \\ iNaturalist2021}& \multirowcell{2}{Acc $\uparrow$ \\ ImageNet}\\  
     \cmidrule{3-4}
     \cmidrule{5-6}
     & & Train $\downarrow$ & Inference $\downarrow$ & Train $\uparrow$& Inference$\uparrow$ & &\\               
     \midrule
    Full fine-tuning                                & 1000 & 8.71 & 4.01 &17.6 & 77.3 &267.4 & 61.1 & 79.8 & 82.5 &89.0 \\ %
    \midrule
    Linear probing                                  & 0     & 4.09 & 4.01&75.9 & 77.6 &267.4 & 57.8 & 73.3 & 70.1 &88.4\\ %
    LST~\cite{sung2022lst}                          & 27 & 4.77 &4.11 &42.5 & 66.4 &274.9 & 59.5 & 75.0 & 76.0 & 88.2 \\  %
    Finetune last layer                             & 50 & 4.30 & 4.01&69.4 & 77.5 &267.4 & 59.5 & 77.4 & 79.2 & 88.8 \\ %
    Prompt Tuning~\cite{jia2022visual}              & 0.6 & 6.00 & 4.04&11.0 & 27.9 &571.8 & 59.2 & 77.5 & 76.8 & 88.3 \\ %
    Finetune last 2 layers                          & 75 & 4.41 & 4.01&64.4 & 77.6 &267.4 & 59.7 & 78.1 & 79.8 & 88.8 \\ %
    Bitfit~\cite{zaken2021bitfit}                   & 0.7 & 4.69 & 4.01&34.2 & 77.5 &267.4 & 58.6 & 78.6 & 76.4 & 88.4 \\ %
    Finetune only mlp~\cite{touvron2022three}                               & 700 & 7.39 & 4.01&23.1 & 77.7 &267.4 & 60.8 & 79.3 & 81.7 & 88.9 \\ %
    Finetune only attention~\cite{touvron2022three}                         & 320 & 5.96 & 4.01&25.7 & 77.4 &267.4 & 60.9 & 79.7 & 81.9 & 88.8\\ %
    LoRA~\cite{hu2021lora}                          & 18 & 4.86 & 4.08 &27.5 & 64.2 &272.2 & 61.5 & 80.9 & 83.3 & 88.6\\ %
    \midrule
    Ours (LoSA)                                     & 4.8 & 4.19 & 4.07 &64.7 & 73.6 &269.4 & 61.3 & 81.3 & 83.8 &89.0 \\ %
    \bottomrule
    \end{tabular}
    }
    \caption{Results on iNaturalist2018/2021, Places365 and ImageNet.
    We used a Vit-g backbone with 1 billion paramaeters for all experiments.
    The number of trainable parameters does not contain the classifier weights, which means that the number of updated parameters is therefore constant across all datasets.
    The speed, memory usage and GFLOPs are barely affected by the number of classes, and we report these metrics for iNaturalist 2018.
    We used data from this table for the Figures 1 and 4 of the main paper, and Fig.~\ref{fig:results_inaturalist2021} and~\ref{fig:results_places365} of this appendix.
    Note that for LoRA~\cite{hu2021lora}, it is possible to absorb the learned weights into the frozen weights after training, to ensure that the inference time and FLOPs does not change at all compared to full finetuning which does not add any parameters to the model~\cite{hu2021lora}. However, we have not implemented this.
    Note that although we average training- and inference-time over 50 batches, there is still some random variation in the timing.
	} 
    \label{tab:main_table}
\end{table*}

Table \ref{tab:main_table} provides detailed results for iNaturalist 2021/2018, Places365 and ImageNet. We also provide inference metrics, such as memory utilization and speed during inference.
Similar to the conclusions from the main paper on iNaturalist2018, our proposed \OURSSHORT is Pareto-optimal on these datasets, as no other method is both more accurate and more efficient (across multiple efficiency metrics).
Figures~\ref{fig:results_inaturalist2021} and~\ref{fig:results_places365} visualise our results on iNaturalist 2021 and Places365 respectively.

As mentioned in the main paper, we have also included results for ImageNet in Tab.~\ref{tab:main_table}.
However, we do not consider ImageNet to be a good dataset for evaluating the performance of adaptation methods, as the performance is saturated, and all methods achieve similar accuracy.
The dataset is likely saturated because it is too similar to the pretraining dataset~\cite{deng_cvpr_2009, sun_iccv_2017, steiner2022train, zhai2022scaling}.
The lowest accuracy achieved is 88.2 for LST~\cite{sung2022lst}, and the highest accuracy is 89.0 for full-finetuning and our method.

Finally, we observe that for inference, the memory consumption and the speed is very similar across all methods.
As described in the main paper, this happens as the efficiency metrics during inference are almost entirely influenced by the size of the backbone and not by the size of the adaptors, which is relatively very small. Therefore, the methods that add additional components (\eg \OURSSHORT, LST~\cite{sung2022lst}, LoRA~\cite{hu2021lora}) behave similarly as the other baselines that do not add any additional components (\eg full finetuning, BitFit~\cite{zaken2021bitfit}) with respect to the efficiency metrics during inference.

\input

\section{Further details and ablation studies about baselines} \label{sec:Implementation_details}

In this section, provide additional details about the baselines used in our main paper.

\paragraph{LoRA~\cite{hu2021lora}}

We implemented LoRA according to the original paper~\cite{hu2021lora} and the official \href{https://github.com/microsoft/LoRA}{repository}.
In the original LoRA paper, the authors only adapted the query- and key-projections within the transformer block.
However, as these experiments were conducted for natural language tasks, we performed further ablations of LoRA in Tab.~\ref{tab:abl_lora_components} and~\ref{tab:ablation_lora_rank}.

Based on our experiments, we found that adapting all linear projections in the transformer block (\ie query-, key-, value- and output-projections, along with the MLP block) performed the best (Tab.~\ref{tab:abl_lora_components}).
Furthermore, we found using a LoRA rank of $r = 32$ to provide a good trade-off between accuracy and efficiency (Tab.~\ref{tab:ablation_lora_rank}).
As a result, we used these settings in all of our experiments in the main paper.

Finally, we note that during inference, it is possible to absorb the learned LoRA parameters back into the original weights of the transformer block~\cite{hu2021lora}.
This means that the inference speed and GFLOPs remains unchanged compared to the backbone.
However, we did not implement this in our experiments.

\begin{table}[t]
\resizebox{\linewidth}{!}{
\begin{tabular}{ccccc cccc}
\toprule
Q & K & V & Out & MLP & Params (M) & GFLOPs & Time (img/sec) & Accuracy \\ \midrule
\checkmark &            &            &            &            & 3.7  & 268.4 & 30.5    &77.7\\    %
\checkmark & \checkmark &            &            &            & 7.4  & 269.4 & 28.9    &77.7\\    %
\checkmark & \checkmark & \checkmark &            &            & 11.1 & 270.3 & 27.3    &79.7\\    %
\checkmark & \checkmark & \checkmark & \checkmark &            & 14.8 & 271.2 & 26.5    &79.9\\    %
\checkmark & \checkmark & \checkmark & \checkmark & \checkmark & 18.5 & 272.2 & 27.5    &80.9\\    %

\bottomrule
\end{tabular}}
\caption{
	Ablation on which components of the transformer block to apply LoRA to, on the iNaturalist2018 dataset, using a Vit-g backbone.
	Our choice of using LoRA on all components provides the best accuracy. 
} 
\label{tab:abl_lora_components}
\end{table}

\begin{table}[t]
\resizebox{\linewidth}{!}{
\begin{tabular}{ccccc}
\toprule
Rank, $r$ & Params (M) & GFLOPs & Time (img/sec) & Accuracy \\ \midrule
1   & 0.58    &267.7&  26.8  & 79.2\\     %
4   & 2.31    &268.1&  27.7  & 79.0\\     %
8   & 4.62    &268.7&  27.8  & 79.9\\     %
16  & 9.24    &269.8&  27.7  & 80.6\\     %
32  & 18.47   &272.2&  27.5  & 80.9\\     %
64  & 36.95   &276.8&  27.3  & 81.1\\     %
128 & 73.89   &286.1&  26.7  & 80.5\\     %
256 & 147.79   &304.8&  25.0  & 79.8\\     %

\bottomrule
\end{tabular}}
\caption{Ablating the LoRA rank on the iNaturalist2018 dataset, using Vit-g backbone. We consider our choice of using $r=32$ a good trade-off.
}
\label{tab:ablation_lora_rank}
\end{table}

\paragraph{BitFit~\cite{zaken2021bitfit}}
For BitFit, we implement it so that it can train every single bias in the whole network. This is in line with the original paper~\cite{zaken2021bitfit} which showed that using all the biases in the network provides the best accuracy. 

\paragraph{Prompt tuning~\cite{jia2022visual}}

We implemented visual prompt-tuning, according to the \href{https://github.com/kmnp/vpt}{authors' implementation} and paper~\cite{jia2022visual}.
We have used the "Deep" prompt-tuning, which means that we have learnable tokens at multiple layers in the architecture, instead of only at the first layer as in "Shallow" prompt-tuning.

We found that prompt-tuning is very sensitive to the number of prompts inserted at each layer~\cite{jia2022visual, hu2021lora, li2021prefix}.
And in some cases, can even perform worse than linear probing.
Therefore, in our experiments, we first did an ablation study to determine a configuration that achieved good accuracy-efficiency trade-offs, as shown in Tab.~\ref{tab:ablation_prompt_tuning}, and used it for future experiments.
Concretely, we used 16 learnable prompts per layer, for the first 24 layers of our ViT-g backbone.

\begin{table}[t]
\resizebox{\linewidth}{!}{
\begin{tabular}{cc ccc}
\toprule
Number of prompts, $P$ & Number of initial layers, $L$ & Params ($10^3$) & GFLOPs & Accuracy \\
\midrule
0   &  0   &  0   & 267.4 & 73.3  \\  %
\midrule

1   &  1   &  1   & 268.5 & 73.4  \\  %
1   &  4   &  6   & 271.6 & 72.9  \\  %
1   &  8   &  11  & 275.3 & 72.7  \\  %
1   &  16  &  23  & 281.4 & 72.8  \\  %
1   &  24  &  34  & 285.9 & 72.4  \\  %
4   &  1   &  6   & 271.7 & 72.5  \\  %
4   &  4   &  23  & 284.0 & 72.6  \\  %
4   &  8   &  45  & 298.9 & 70.9  \\  %
4   &  16  &  90  & 323.6 & 72.4  \\  %
4   &  24  &  135 & 341.6 & 75.4  \\  %
8   &  1   &  11  & 276.0 & 72.7  \\  %
8   &  4   &  45  & 300.6 & 71.4  \\  %
8   &  8   &  90  & 330.5 & 70.5  \\  %
8   &  16  &  180 & 380.5 & 73.8  \\  %
8   &  24  &  270 & 417.0 & 77.1  \\  %
16  &  1   &  23  & 284.6 & 72.3  \\  %
16  &  4   &  90  & 334.0 & 70.6  \\  %
16  &  8   &  180 & 394.5 & 70.8  \\  %
16  &  16  &  361 & 496.4 & 75.1  \\  %
16  &  24  &  541 & 571.8 & 77.5  \\  %
24  &  1   &  34  & 293.3 & 72.4  \\  %
24  &  4   &  135 & 376.6 & 70.1  \\  %
24  &  8   &  270 & 459.2 & 73.0  \\  %
24  &  16  &  541 & 615.0 & 74.7  \\  %
24  &  24  &  811 & 731.6 & 78.4  \\  %

\bottomrule
\end{tabular}}
\caption{Ablating prompt tuning on the iNaturalist2018 dataset with a Vit-g backbone.
Prompt tuning is sensitive to the number of learned prompts, $P$, and the number of initial layers, $L$, which these prompts are added to.
(Note that prompts are not added only to the input, but the initial $L$ layers in~\cite{jia2022visual}).
$P = 0, L = 0$, corresponds to a linear probing baseline.
Based on these results, we used $P = 16$, $L = 24$.
}
\label{tab:ablation_prompt_tuning}
\end{table}

\paragraph{LST~\cite{sung2022lst}}

Our implementation is based on the \href{https://github.com/ylsung/Ladder-Side-Tuning/blob/main/VL-T5/src/my_transformers/modeling_bart.py}{public code} and paper~\cite{sung2022lst}.
The paper uses a transformer model as the parallel network.
Following the original authors, we used a linear projection to reduce the dimensionality of the activations from the backbone, to the hidden dimension, $d$, of the parallel network.
Following the authors, we used $d = 48$.

\paragraph{ST-Adapter~\cite{pan2022st}}
We have followed the \href{https://github.com/linziyi96/st-adapter}{public implementation.} and paper~\cite{pan2022st}.
In the paper, the authors provide multiple ways of inserting the ST-Adapter into the backbone architecture.
We used the variant which achieves the best accuracy, namly inserting an ST-Adapter module before and after the Multihead Self-Attention (MHSA) in each transformer block.
Following the paper, our convolutional kernel size is 3x3x3, and the hidden dimension of the ST-Adapter is 768.

\section{VTAB} \label{sec:VTAB}
Table~\ref{tab:vtab_supplementary} compares to additional methods on the VTAB-1K benchmark (in the main paper, we presented fewer methods due to space constraints).
Our LoSA method still achieves superior accuracy-efficiency trade-offs compared to prior work.

\begin{table*}[t]

\centering
\setlength{\tabcolsep}{0.3pt}
\setlength{\tabcolsep}{1pt}
\scalebox{0.82}{
\begin{tabular}{p{2.9cm} P{0.69cm}|P{0.69cm}P{0.69cm}P{0.69cm}P{0.69cm}P{0.69cm}P{0.69cm}P{0.69cm}|P{0.69cm}P{0.69cm}P{0.69cm}P{0.69cm}|P{0.69cm}P{0.69cm}P{0.69cm}P{0.69cm}P{0.69cm}P{0.69cm}P{0.69cm}P{0.69cm}|P{0.69cm}P{0.69cm}P{0.69cm}P{0.69cm}}
	\toprule[1.5pt]
	\multicolumn{2}{c|}{}&\multicolumn{7}{c|}{\textbf{Natural}}&\multicolumn{4}{c|}{\textbf{Specialized}}&\multicolumn{8}{c|}{\textbf{Structured}}&\\
	&\multicolumn{1}{c|}{\rotatebox[origin=l]{90}{Param $\downarrow$ ($10^6$)}}
	&\multicolumn{1}{c}{\rotatebox[origin=l]{90}{Cifar100}}
	&\multicolumn{1}{c}{\rotatebox[origin=l]{90}{Caltech101}}
	&\multicolumn{1}{c}{\rotatebox[origin=l]{90}{DTD}}
	&\multicolumn{1}{c}{\rotatebox[origin=l]{90}{Flower102}}
	&\multicolumn{1}{c}{\rotatebox[origin=l]{90}{Pets}}
	&\multicolumn{1}{c}{\rotatebox[origin=l]{90}{SVHN}}
	&\multicolumn{1}{c|}{\rotatebox[origin=l]{90}{Sun397}}
	&\multicolumn{1}{c}{\rotatebox[origin=l]{90}{Camelyon}}
	&\multicolumn{1}{c}{\rotatebox[origin=l]{90}{EuroSAT}}
	&\multicolumn{1}{c}{\rotatebox[origin=l]{90}{Resisc45}}
	&\multicolumn{1}{c|}{\rotatebox[origin=l]{90}{Retinopathy}}
	&\multicolumn{1}{c}{\rotatebox[origin=l]{90}{Clevr-Count}}
	&\multicolumn{1}{c}{\rotatebox[origin=l]{90}{Clevr-Dist}}
	&\multicolumn{1}{c}{\rotatebox[origin=l]{90}{DMLab}}
	&\multicolumn{1}{c}{\rotatebox[origin=l]{90}{KITTI-Dist}}
	&\multicolumn{1}{c}{\rotatebox[origin=l]{90}{dSpr-Loc}}
	&\multicolumn{1}{c}{\rotatebox[origin=l]{90}{dSpr-Ori}}
	&\multicolumn{1}{c}{\rotatebox[origin=l]{90}{sNORB-Azim}}
	&\multicolumn{1}{c|}{\rotatebox[origin=l]{90}{sNORB-Ele}}
	&\multicolumn{1}{c}{\rotatebox[origin=l]{90}{Avg Natural}}
	&\multicolumn{1}{c}{\rotatebox[origin=l]{90}{Avg Specialized}}
	&\multicolumn{1}{c}{\rotatebox[origin=l]{90}{Avg Structured}}
	&\multicolumn{1}{c}{\rotatebox[origin=l]{90}{Average}}\\
	\specialrule{0em}{1pt}{1pt}
	\hline
	\specialrule{0em}{1pt}{1pt}
	\multicolumn{22}{l}{\emph{Traditional Finetuning}}\\
	\hline
	\specialrule{0em}{1pt}{1pt}
	Full\cite{jia2022visual,jie2022convolutional} &85.8&68.9&87.7&64.3&97.2&86.9&87.4&38.8&79.7&95.7&84.2&73.9&56.3&58.6&41.7&65.5&57.5&46.7&25.7&29.1&75.9&83.4&47.6&68.9 \\
	Linear\cite{jia2022visual,jie2022convolutional}&0&64.4&85.0&63.2&97.0&86.3&36.6&51.0&78.5&87.5&68.5&74.0&34.3&30.6&33.2&55.4&12.5&20.0&9.6&19.2&69.1&77.1&26.9&57.6\\
	\hline
	\specialrule{0em}{1pt}{1pt}
	\multicolumn{22}{l}{\emph{Efficient adaptation methods}}\\
	\hline
	\specialrule{0em}{1pt}{1pt}
	BitFit\cite{zaken2021bitfit,jie2023fact}&0.10&72.8&87.0&59.2&97.5&85.3&59.9&51.4&78.7&91.6&72.9&69.8&61.5&55.6&32.4&55.9&66.6&40.0&15.7&25.1&73.3&78.3&44.1&65.2\\
    VPT-Shal.\cite{jia2022visual,jie2023fact}&\underline{0.06}&77.7&86.9&62.6&97.5&87.3&74.5&51.2&78.2&92.0&75.6&72.9&50.5&58.6&40.5&67.1&68.7&36.1&20.2&34.1&76.8&79.7&47.0&67.8 \\
	VPT-Deep\cite{jia2022visual,jie2022convolutional}&0.53&78.8&90.8&65.8&98.0&88.3&78.1&49.6&81.8&96.1&83.4&68.4&68.5&60.0&46.5&72.8&73.6&47.9&32.9&37.8&78.5&82.4&55.0&72.0 \\

    RS-Bypass.~\cite{jiang2023restuning}&0.42&64.5&88.8&73.2&99.4&90.6&63.5&\underline{57.2}&85.5&95.2&82.4&75.2&70.4&61.0&40.2&66.8&79.2&52.6&26.0&\underline{49.3}&76.7&84.6&55.7&72.3 \\
    
    Express~\cite{das2023learning}& 0.2*&78.0&89.6&68.8&98.7&88.9&81.9&51.9&84.8&96.2&80.9&74.2&66.5&60.4&46.5&77.6&78.0&49.5&26.1&35.3&79.7&84.0&55.0&72.9 \\
    TOAST~\cite{shi2023toast}&14.0&82.1&90.5&70.5&98.7&89.7&71.9&53.3&84.3&95.5&85.5&74.2&75.4&60.8&44.7&77.5&73.9&47.5&24.5&33.7&79.5&84.9&54.8&73.1 \\
 
    Adapter \cite{houlsby2019parameter, jie2022convolutional}&0.16&69.2&90.1&68.0&98.8&89.9&82.8&54.3&84.0&94.9&81.9&75.5&80.9&65.3&48.6&78.3&74.8&48.5&29.9&41.6&79.0&84.1&58.5&73.9 \\
	LoRA\cite{hu2021lora,jie2022convolutional}&0.29&67.1&91.4&69.4&98.8&90.4&85.3&54.0&84.9&95.3&84.4&73.6&\underline{82.9}&\bf69.2&49.8&78.5&75.7&47.1&31.0&44.0&79.5&84.6&59.8&74.5
	\\
	AdaptFormer\cite{chen2022adaptformer,jie2022convolutional}&0.16&70.8&91.2&70.5&99.1&90.9&86.6&54.8&83.0&95.8&84.4&76.3&81.9&64.3&49.3&80.3&76.3&45.7&31.7&41.1&80.6&84.9&58.8&74.7 \\
    NOAH \cite{zhang2022neural, jie2022convolutional}&0.36&69.6&92.7&70.2&99.1&90.4&86.1&53.7&84.4&95.4&83.9&75.8&82.8&\underline{68.9}&49.9&81.7&81.8&48.3&32.8&44.2&80.3&84.9&61.3&75.5\\
    FacT-TK \cite{jie2023fact}& \underline{0.06}&70.6&90.6&70.8&99.1&90.7&88.6&54.1&84.8&96.2&84.5&75.7&82.6&68.2&49.8&80.7&80.8&47.4&33.2&43.0&80.6&85.3&60.7&75.6\\
	SSF \cite{lian2022scaling} &0.24&69.0&92.6&75.1&99.4&\underline{91.8}&90.2&52.9&\bf87.4&95.9&\underline{87.4}&75.5&75.9&62.3&\bf53.3&80.6&77.3&54.9&29.5&37.9&81.6&86.6&59.0&75.7\\
	 Convpass\textsubscript{attn} \cite{jie2022convolutional}&0.16&71.8&90.7&72.0&99.1&91.0&89.9&54.2&85.2&95.6&83.4&74.8&79.9&67.0&50.3&79.9&84.3&53.2&34.8&43.0&81.2&84.8&61.6&75.8\\
    RepAdapter
     \cite{luo2023towards}&0.22&72.4&91.6&71.0&99.2&91.4&90.7&55.1&85.3&95.9&84.6&75.9&82.3&68.0&50.4&79.9&80.4&49.2&38.6&41.0&81.6&85.4&61.2&76.1\\
     RS~\cite{jiang2023restuning}&0.55&75.2&92.7&71.9&99.3&\bf{91.9}&86.7&\bf{58.5}&86.7&95.6&85.0&74.6&80.2&63.6&50.6&80.2&85.4&55.7&31.9&42.0&82.3&85.5&61.2&76.3 \\
	Convpass \cite{jie2022convolutional}&0.33&72.3&91.2&72.2&99.2&90.9&\bf91.3&54.9&84.2&96.1&85.3&75.6&82.3&67.9&51.3&80.0&85.9&53.1&36.4&44.4&81.7&85.3&62.7&76.6\\
 HST~\cite{lin2023hierarchical}&0.78&76.7&\bf94.1&74.8&\underline{99.6}&91.1&\underline{91.2}&52.3&\underline{87.1}&96.3&\bf88.6&\underline{76.5}&\bf85.4&63.7&\bf{52.9}&81.7&87.2&56.8&35.8&\bf52.1&\bf82.8&\bf87.1&64.5&78.1 \\
	\hline
	\OURSSHORT, $r = 16$ &0.19&\underline{82.5}&92.8&76.1&\bf99.7&90.5&82.0&55.8&86.6&\bf97.1&87.0&\bf76.7&81.5&62.3&48.6&82.1&\bf94.2&\bf61.7&\underline{47.9}&45.6&\bf82.8&86.9&\bf65.5&\bf78.4\\ %
	\OURSSHORT, $r = 8$ &0.10&82.2&92.7&\bf76.7&\bf99.7&90.7&81.0&55.4&86.9&\bf97.1&\underline{87.4}&\underline{76.5}&79.9&61.8&48.6&\underline{82.4}&\underline{92.3}&\underline{61.1}&\bf48.7&47.3&\underline{82.6}&\underline{87.0}&\underline{65.3}&\underline{78.3}\\ %
	
	\OURSSHORT, $r = 4$ &\bf{0.05}&\bf82.7&\underline{93.0}&\underline{76.2}&\bf99.7&89.8&80.0&56.1&86.3&\underline{96.7}&86.7&76.3&78.8&61.4&48.0&\bf82.6&91.7&58.4&46.9&47.6&82.5&86.5&64.4&77.8\\ %

	\bottomrule[1.5pt]

\end{tabular}
}
\vspace{-0.75\baselineskip}
\caption{
Comparison to state-of-the-art parameter-efficient finetuning methods on VTAB-1K~\cite{zhai2019large}.
Following standard practice, the final ``Average'' is the average of three preceding groupwise averages.
Parameters denotes the number of learnable parameters excluding the final classification layer, as the number of parameters in this final layer depend on the number of classes, which varies between 2 and 397.
Each variant of our model, which we obtain by varying the rank $r$ of our Low-rank Mixer Block, achieves better accuracy-parameter trade-offs than previous approaches, when using the same ViT-B backbone.
\textbf{Best results} are bolded, and \underline{second-best} underlined. * denotes that we estimated the number of learnable parameters in Express~\cite{das2023learning} based on the hyperparameters presented in the paper, as the authors did not explicitly state the number of learned parameters in their paper.
}
\label{tab:vtab_supplementary}
\end{table*}

\section{Implementation details} \label{sec:Hyperparam}

In this section we provide more details for the hyperparameters that we have used for our methods on the large scale datasets (image and video domains).
Table~\ref{tab:hyperparam_images} shows that we use the same hyperparameters across different image datasets, and achieve strong results across all of them.
Table~\ref{tab:hyperparam_videos} presents our hyperparameters for video classification.

Our training hyperparameters for image classification are based on those from~\cite{steiner2022train, zhai2022scaling}, whilst our training hyperparameters for video classification are based on those from~\cite{arnab2021vivit}.
Overall, we did not change hyperparametrers of our model (such as the rank $r$ of \OURSSHORT) between images and video.

\begin{table}[t]
    \centering
    \resizebox{\linewidth}{!}{
    \begin{tabular}{lcccc}
    \toprule
     \multirowcell{2}[0pt][l]{Hyperparameter} &  \multirowcell{2}{iNaturalist2018} & \multirowcell{2}{iNaturalist2021} & \multirowcell{2}{Places365}& \multirowcell{2}{ImageNet}\\  
    & & &  \\               
    \hline
    Optimiser                                & \multicolumn{4}{c}{Momentum, $\lambda = 0.9$} \\
    Batch size                               & \multicolumn{4}{c}{512} \\
    Learning rate scheduler                  & \multicolumn{4}{c}{cosine} \\
    Linear warmup steps                      & \multicolumn{4}{c}{500} \\
    Base learning rate                       & \multicolumn{4}{c}{0.05} \\
    Number of training steps                 &  \multicolumn{4}{c}{20 000} \\
    Rank, $r$                                &  \multicolumn{4}{c}{64} \\
    Input resolution                         & \multicolumn{4}{c}{224} \\
    \bottomrule
    \end{tabular}
    }
    \caption{LoSA hyperparameters for large-scale image classification datasets.}
    \label{tab:hyperparam_images}
\end{table}
\begin{table}[t]
    \centering
    \small{
        \begin{tabular}{lc}
        \toprule
        Hyperparameter & Kinetics 400 \\ 
        \midrule
        Optimiser & momentum, $\lambda = 0.9$ \\
        Batch size & 64 \\
        Learning rate scheduler & cosine\\
        Linear warmup epochs &  2.5 \\
        Base learning rate & 0.04 \\
        Epochs & 30 \\
        Rank, $r$ & 64\\
        Input resolution & 224\\
        \bottomrule
        \end{tabular}
        \caption{LoSA hyperparameters for video classification task.
        }
    \label{tab:hyperparam_videos}
    }
\end{table}

We used synchronous SGD with distributed data-parallel training.
For images, we used a local batch size of 16.
For video, our local batch size was 1 for all experiments in the main paper. 

When we report the memory consumption for training or inference (Fig.~1 and 4, Tabs.~2, 3, 4, 5 of the main paper, and Fig.~\ref{fig:results_inaturalist2021} and~\ref{fig:results_places365} and Tab.~\ref{tab:main_table} of the appendix, we report the memory usage with a local batch size of 1.
The reason for that is that it removes the effect of the batch size (which is a training hyperparameter), and it also has a clear intepretation: it represents the minimum possible memory that we require during data-parallel training.

When reporting the training and inference speeds, we average the runtime over 50 batches.

We implemented our method and baselines using the Scenic library~\cite{dehghani2021scenic} and JAX~\cite{jax2018github}.

\end{document}